\documentclass[onecolumn]{article}
\usepackage{spconf,amsmath,epsfig}
\usepackage{times}
\usepackage{epsfig}
\usepackage{graphicx}
\usepackage{amsmath}
\usepackage{amssymb}
\usepackage{subfigure}
\usepackage{cite}
\usepackage{color}
\usepackage{multirow}
\usepackage{epstopdf}
\usepackage{multicol}
\usepackage{multirow}

\begin{document}\sloppy

\def\x{{\mathbf x}}
\def\L{{\cal L}}

\title{Deep attention-guided fusion network for lesion segmentation}
%
\name{Hengliang Zhu$^1$, Yangyang Hao$^1$, Lizhuang Ma$^1$, Ruixing Li$^2$, Hua Wang$^2$}
\address{Dept. of Computer Science and Engineering, Shanghai Jiao Tong University, China \\
$^2$Dept. of Computer Science, Minjiang teachers College, China}

\maketitle
\begin{abstract}
We participated the Task 1: Lesion Segmentation. The paper describes our algorithm and the final result of validation set for the ISIC Challenge 2018 - Skin Lesion Analysis Towards Melanoma Detection.
\end{abstract}
\section{Introduction}
The skin lesion segmentation has attracted many researchers in recent years. In many medical images processing tasks, lesion segmentation is usually utilized as a pre-precessing step, such as lesion attribute detection, lesion Diagnosis. However, it is challenging to automatically detect the lesion region within dermoscopic images. The lesion region in the images usually contain different patterns, such as color, size, shape and position, so the detection model should be able to efficiently obtain the high-level features of target object.
Saliency detection is also a hot topic in recent years. From the perspective of biological visual perception, salient object has abundantly visual information, which is first attracted human attentions. Therefore, saliency detection aims at locating the most visually distinctive regions in an image.

Inspired by this mechanism, we propose a novel attention-guided fusion network (see Fig. 1) to detect the skin lesion region. The proposed model is also based on a fully convolutional neural network~\cite{long2015fully,Zhang2017Amulet} with the end-to-end manner. In our network, a simple yet efficient fusion method is used to utilize the multi-level features. The deep model can locate the lesion melanoma and generate the highlighted segmentation map. Experiments indicate that the training process is efficient and the model has the capacity of detecting the melanoma.

\section{Overview of Architecture}
The architecture of our deep attention-guided network consists of two components, a multi-path feature fusion framework and a coupled structure module (CSM). The multi-path fusion is a spatial pyramid structure, which is used to produce different semantic feature maps. For each path, we use a simple way to integrate its feature map with shallow convolutional layer. At the same time, we propose a coupled structure module to improve the accuracy of localization for saliency detection. Finally, the final lesion segmentation result is generated by aggregating maps from different paths.
\begin{figure*}[htb]
  \centering
  \includegraphics[scale=0.5]{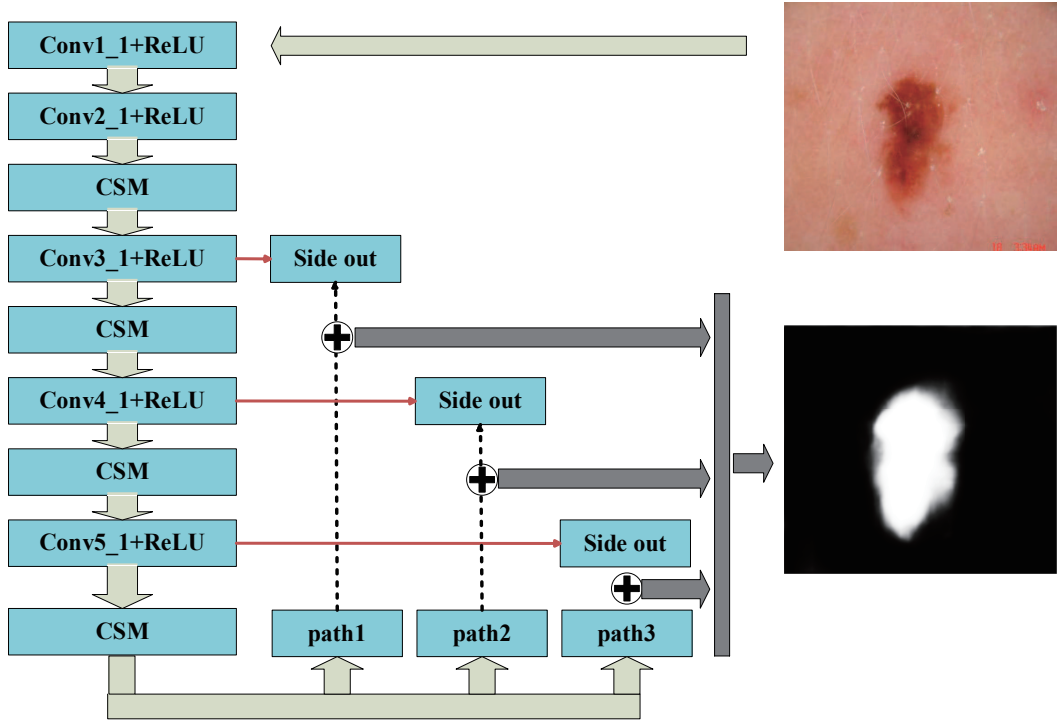}
  \caption{The architecture of our attention-guided model. The proposed end-to-end network is also built on VGG16 model. Given an original image, the high-level feature of lesion region is first extracted by each path. Then three fusion maps are obtained by integrating these features. At last, the final segmentation result is generated by merging the three feature maps.} \label{fig:framework}
\end{figure*}

\begin{figure}[htb]
  \centering
  \includegraphics[scale=0.6]{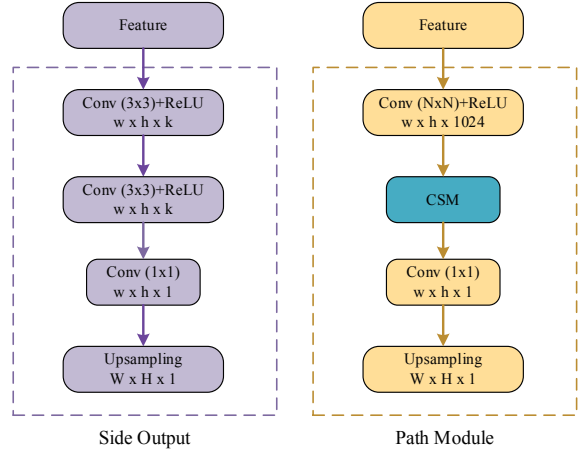}
  \caption{The structure of side output and path module.} \label{fig:pathModule}
\end{figure}

\begin{figure}[htb]
  \centering
  \includegraphics[scale=0.5]{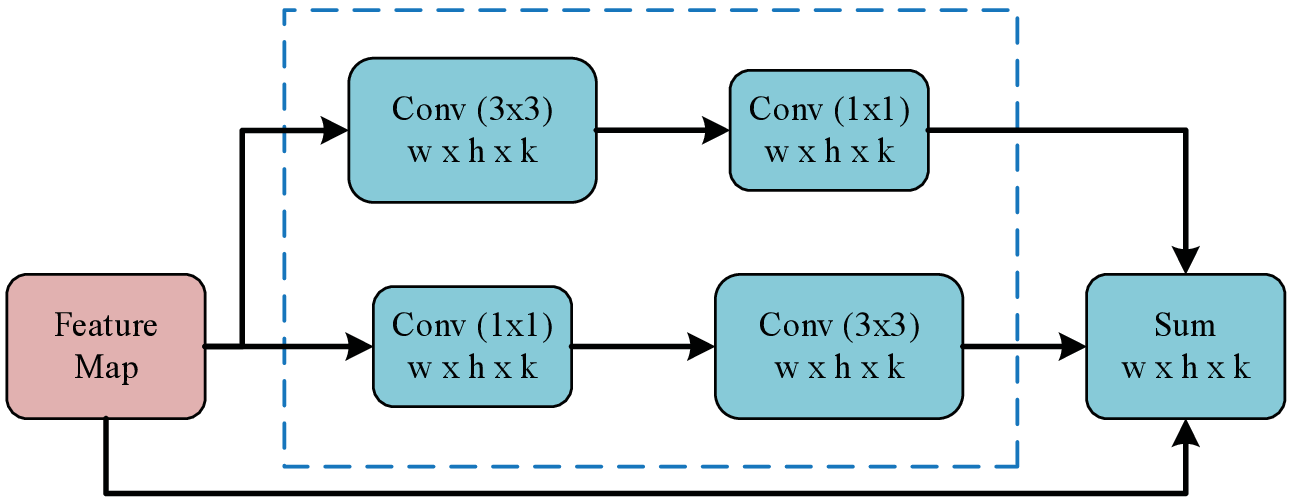}
  \caption{The architecture of CSM.} \label{fig:coupledStructure}
\end{figure}

\textbf{Network framework.}  Our skin lesion detection algorithm is based on the VGG16 model, which has a strong feature representation ability. The proposed network has five max-pooling layers with kernel size 2 and stride 2, so the size of output feature map reduced by a factor of 32. To make the output map have the same size as input image, we use up-sampling operation to scale the map in the network. In order to improve the accuracy of saliency detection, we add three paths after the last pooling stage to build the spatial pyramid structure. We connect three extra branches to shallow convolutional layers. These shallow layers are conv3-1, conv4-1 and conv5-1, respectively. Each branch contains two $3 \times 3$ convolutional layers, one $1 \times 1$ convolutional layer and one up-sampling layer. Then, three output segmentation results (Side Output, see Fig. 2) that represent rich contextual information are generated by each branch, denoted as $B_{i}$. These hierarchical segmentation results $\{B_{i}, i = 1, 2, 3\}$ contain abundantly low-level visual properties that is complementary to the high-level segmentation results $\{P_{i}, i = 1, 2, 3\}$. Thus, we further fuse these segmentation results, defined as
\begin{equation}\label{fusion}
F_{i} = W_{i} * Concat(P_{i},B_{i}),
\end{equation}
where the symbol $*$ is convolution operation; $Concat$ is the cross-channel concatenation. $W_{i}$ is a convolutional layer with $1\times1$ kernel size, which is used to balance the importance of each feature map.

We use the cross entropy in our model, and compute the loss function between the fused segmentation result and the ground truth. Given a input image $X$, we define its corresponding segmentation as $Y=\{y_{i},i=1,...,|Y|\},y_{i}\in[0,1]$. Thus, the final loss function can be represented by
\begin{equation}\label{loss}
\begin{split}
L_{final}=&-\sum_{y_{i} \in Z}y_{i}logP(y_{i}=1|X;\Phi)\\
&+(1-y_{i})logP(y_{i}=0|X;\Phi).
\end{split}
\end{equation}
The parameter $\Phi$ is the collection of all network weights, which are updated using SGD algorithm at each iteration.

\textbf{Coupled structure.}  The coupled structure (see Fig. 3) consists of two complementary and symmetric components. Each component includes one $3 \times 3$ convolutional layer and one $1 \times 1$ convolutional layer. The non-linear transformation (ReLU) is also used after each convolutional layer. Besides, in order to enlarge the receptive filed to cover the whole object, we use the dilated convolutions to increase the size of filters.

\subsection{Pre-processing}
We use RGB images as input and the training images are resized to $224 \times 224$. For testing stage, the validation set and testing set are also resized to $224 \times 224$. At last, the segmentation map are recovered to the original size of the images.
\subsection{Training}
We train our skin lesion detection model by using an open source deep learning framework Caffe~\cite{Jia2014Caffe}, and directly feed the input images into the network. For training stage, it takes us about 8 hours to train the deep model. For testing stage, our deep model takes about 0.03s (33 FPS) to process an image without any pre/post-processing.

\subsection{Post-processing}
To further make the segmentation smoothness, we use the fully connected CRF method~\cite{Kr2012Efficient} to optimize the segmentation result. We use the posterior probability of each pixel being salient as the final segmentation result, defined as follows:
\begin{equation}\label{equ:pairwise}
\begin{split}
\delta_{i,j}(y_{i},y_{j})=&\mu(y_{i},y_{j})[\omega_{1}exp(-\frac{\|p_{i}-p_{j}\|^{2}}{2\sigma_{\alpha}^{2}} \\ &-\frac{\|I_{i}-I_{j}\|^{2}}{2\sigma_{\beta}^{2}})+\omega_{2}exp(-\frac{\|p_{i}-p_{j}\|^{2}}{2\sigma_{\gamma}^{2}})],
\end{split}
\end{equation}
The details of the parameters can be found in work~\cite{Kr2012Efficient}. Note that the CRF refinement is only used in inference phase. Then, an adaptive threshold method is used to segment the image into a binary image. After morphological dilation and morphological corrosion, small holes are filled. The official description states that the result has only one object, so the small regions that far away from the center of the image are classified as background.

\subsection{Datasets}
We participated the part I of ISBI Challenge 2018 - Skin Lesion Analysis Towards Melanoma
Detection: Lesion Segmentation, featuring image contour detection. The training data consists of 2594 images and 2594 corresponding ground truth response masks. The data was extracted from the "ISIC 2018: Skin Lesion Analysis Towards Melanoma Detection" grand challenge datasets~\cite{Codella2016Skin,Tschandl2018The}. The lesion images were acquired with a variety of dermatoscope types, from all anatomic sites (excluding mucosa and nails), from a historical sample of patients presented for skin cancer screening, from several different institutions. Every lesion image contains exactly one primary lesion; other fiducial markers, smaller secondary lesions, or other pigmented regions may be neglected. The validation data consists of 100 images. The test data consists of 1000 images and the final rank is based on Jaccard index.

\subsection{Implementation}
The proposed algorithm is trained on a Intel Core computer with an i7-6850K CPU and a single GeForce GTX 1080Ti GPU. In our experiments, we utilize VGG16 as our pre-trained model and set the base learning rate to $1e-8$. The parameter of momentum is set to 0.9 and the weight decay is set to 0.0005. In our experiments, we set the parameters of $\omega_{1}$, $\omega_{2}$, $\sigma_{\alpha}$, $\sigma_{\beta}$ and $\sigma_{\gamma}$ are set to 3.0, 5.0, 3.0, 60.0 and 3.0 respectively.

\subsection{Results}
Our method yielded an average Jaccard index of 0.78 on the online validation dataset.

\bibliographystyle{IEEEbib}
\bibliography{isic2018template}

\end{document}